\documentclass[11pt]{article}
\usepackage[hyperref]{ccl2020-en}
\usepackage{times}
\usepackage{url}
\usepackage{CJKutf8}
\usepackage{latexsym}
\usepackage{graphicx}
\usepackage{fancyhdr}
\usepackage{verbatim}
\usepackage{times}
\usepackage{multirow}
\usepackage{url}

\title{LiveQA: A Question Answering Dataset over Sports Live}

\author{Qianying Liu\thanks{\quad This denotes equal contribution.}$\ \,^{12}$, Sicong Jiang\footnotemark[1]$\ \,^{1}$, Yizhong Wang$^{13}$ and Sujian Li$^1$\\
$^1$ Key Laboratory of Computational Linguistics, MOE, Peking University \\
$^2$ Graduate School of Informatics, Kyoto University \\
$^3$ 	University of Washington\\
  {\tt ying@nlp.ist.i.kyoto-u.ac.jp; 512580728@qq.com;}\\{\tt  yizhongw@cs.washington.edu; lisujian@pku.edu.cn} \\
  }

\date{}

\begin{document}
\maketitle
\begin{abstract}
In this paper, we introduce LiveQA, a new question answering dataset constructed from play-by-play live broadcast. It contains 117k multiple-choice questions written by human commentators for over 1,670 NBA games, which are collected from the Chinese Hupu\footnote{https://nba.hupu.com/games} website. Derived from the characteristics of sports games, %LiveQA requires a system to perform reasoning across timeline-based live broadcast, 
LiveQA can potentially  test the  reasoning ability across timeline-based live broadcasts, 
which is challenging compared to the existing datasets.
%which introduces various challenges not found in existing datasets: 
In LiveQA, the questions require understanding the timeline, tracking events or doing mathematical computations. 
%Such abilities are necessary for an intelligent agent and may drive the next-generation question answering system. 
Our preliminary experiments show that the dataset introduces a challenging problem for question answering models, and  a strong baseline model only achieves the accuracy of 53.1\%  and cannot beat the dominant option rule. 
We release the code and data of this paper for future research.\footnote{code: https://github.com/PKU-TANGENT/GAReader-LiveQA}\footnote{data: https://github.com/PKU-TANGENT/LiveQA}
\end{abstract}

\section{Introduction}

The research of question answering (QA), where a system needs to understand a piece of reading material and answer corresponding questions, has drawn considerable attention in recent years. While various QA datasets have been constructed to study how a QA system can understand a specific passage, the common sense knowledge and so on~\cite{rajpurkar2016squad,lai2017race,dunn2017searchqa,rajpurkar2018know}, %most answers of these datasets could be extracted from a few relevant sentences
most questions in these datasets could be given their answers by extracting from a few relevant sentences so that the model only needs to find a small set of supporting evidences,% to infer the answers, where the temporal order of these information does not effect the final answer. 
whose temporal ordering  does not effect the final answer. 
In other words, these questions are raised only considering a fixed document. However, in the real-life question answering, a question could have its \textbf{timelines}. To  infer the answer, a good model needs to understand series of timeline information. For example, the question ``how many points did Lebron James have?'' would have different answers based on the time when the question was asked during a basketball game, and the answer would continuously change during the game. The other question ``Which team would first earn 10 points?'' would require a system to track down information of scoring points along the timeline until one team achieves 10 points.
%Various types of QA datasets have been constructed \cite{rajpurkar2016squad,lai2017race, dunn2017searchqa, rajpurkar2018know}, 

%Question-answering datasets are drawing increasingly attention in the past few years(e.g.,\citet{rajpurkar2016squad} \citet{rajpurkar2018know}). 
% Samples of most existing datasets are crowd-sourced(\cite{lai2017race}) or generated using a search engine (\cite{dunn2017searchqa}).
%without much consideration of 
%, and almost all of them can be separated from their context. 
% It's a common method to find the most relevant sentences from the whole passage and then extract a text span as the final answer (e.g.,~\cite{clark2018think},~\cite{rajpurkar2018know}). 
%In the existing datasets, a question is raised without much consideration of the position where the answer is located.
%In other words, at which part of the passage a question is raised is not necessary to its solution. 
%However, in the real world, the same question may have several correct answers that are closely related to the time when the question is raised.
%is usually produced based on a specific context.
%For example, the question ``how many points did Lebron James have?'' will have different answers based on the question time.
%sometimes the answer to one question is closely related to 
%We consider it as a gap which has not been covered by existing datasets. %What's more, current datasets test little about skills to gain  scattered information. 
According to the analysis above, we consider the timeline-based question answering problem as a gap  which has not been covered by existing datasets.
Thus, in this work we hope to construct a dataset where passages and questions both have timelines and question respondents are required to judge what information should be gathered for the questions involved in a timeline.
%information over the passage. 
%In addition,  questions should lie in the passage and associate with the timeline closely. 
Such a timeline inference-involved QA dataset introduces a new research line of reading comprehension, that evaluates the ability of understanding temporal information of a QA model.
%will help enhance the reading comprehension ability of machine learning systems in general.
%Some of them requires logical reasoning, which means it's hard to answer the questions through comparisons between candidate answers and questions or passages on word-level. However, there's not a considerable amount of this type of datesets. And what's more, the best methods to solve  some of the questions turn out to be matching between the elements of sentences. Thus, the purpose of creating the datasets is not realized. 
\begin{figure}[!t]                                                     %开始插图
\centering                                                                %图片居中
\includegraphics[scale=0.45]{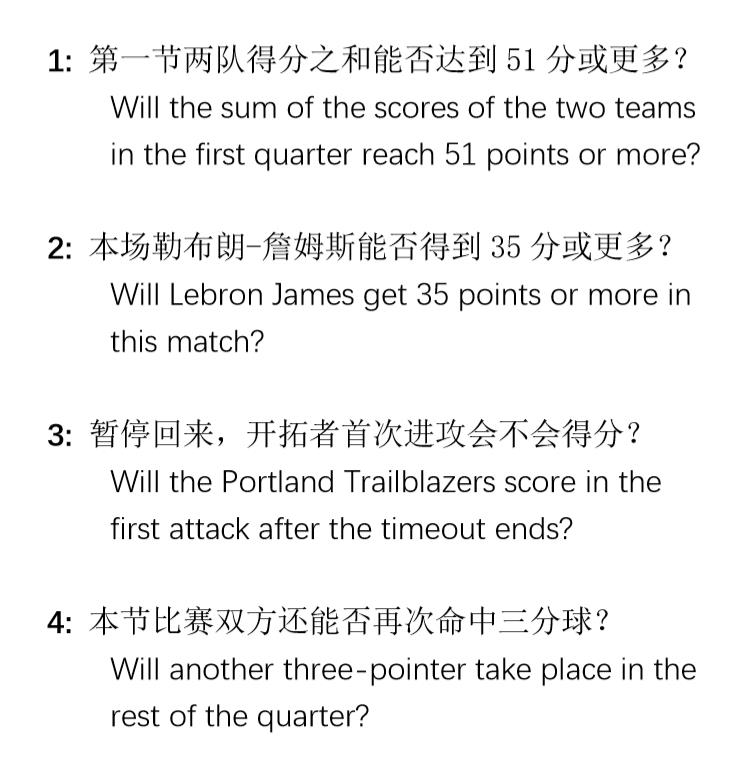}                                    %确定图片大小/位置
\caption{Question Examples from the LiveQA dataset.}                                    %加名称和标签
\end{figure}

Additionally, the real-world questions are often involved in some math calculation, such as addition, subtraction and counting.
To answer the questions correctly, one not only needs to  locate  some specific sentences, but also do calculation or comparison on the extracted evidence. For example, ``How many points did the winner team win?'' needs one system to perform subtraction on the final score to get the correct answer.

To these ends, we construct a QA dataset \textit{LiveQA} based on a Hupu-live-broadcasting-dataset, which is a set of Chinese live-broadcasting passages of NBA. Hupu is a sports news website that has live-broadcasting for basketball games.
%Differing from the play-by-plays in English media, the Hupu passages are written by real humans, while the play-by plays are carrying out automatically. 
In the Hupu-live-broadcasting, the host of one sport game describes the details of the game vividly with emotion and different sentence structures, and presents many game-related quizzes during the game.
We collect the description texts and their quizzes into  LiveQA.
%A host of the game will describe the details of the game vividly with emotion and different sentence structures. What makes it more attractive is the host will present many prize-giving quiz during the game. 
%The quizzes are usually released before an event takes place, but 
Answering the quizzes requires one model to correctly understand the timeline information of the context: some quizzes ask about information of one-whole quarter of the game or which player reaches a certain score earlier. Thus, the model needs to fully understand the temporal information of the live-broadcasting and then performs inference based on the temporal information. 
%Most of the quizzes may be answered correctly by examining their context. Some challenging quizzes require all information in a quarter, while some quizzes are able to be answered through reading the next two sentences. The text range needs to be judged and conducted inference first when answering the question.
Figure 1 shows four question examples in the LiveQA dataset. Answering the first two questions requires an addition math operation, and the 3$^{rd}$ and 4$^{th}$ questions need comparison operation. Meanwhile, we can see that all these questions are time-dependent and require temporal inference.

In summarize, the main characteristics of our LiveQA dataset include the following two aspects.
Firstly,  the questions are time-awared. The model needs temporal inference to obtain the final answer.
% Ingoring such information would lead to a different answer 
%and may match different answers located in different parts of a text without considering its time. 
Secondly, in our dataset, reading comprehension is not limited to extracting a few specific text spans from the document,
%match the question and text, 
but is involved with math calculation.
These characteristics make LiveQA challenging for previous QA systems to answer its questions. 
In this paper, we present an analysis of the resulting dataset to show how these characteristics appear in the data. We also show how questions are involved with temporal inference, and these questions also require mathematical inference. To demonstrate how these characteristics affect the performance of the QA model, we design a pipeline method, which first tries to find supporting sentences and then uses a strong baseline multi-hop inference model named Gated-Attention Reader, to judge the baseline performance on LiveQA. Our experimental results show that 
%the dataset is a challenging problem for machine comprehension models, that 
such strong baseline model only slightly exceeds random choice, which achieve 53.1\% and cannot beat the dominant option rule. 
The analysis and experimental results show how this dataset can effectively examine how a QA system can perform multi-hop temporal and mathematical inference, which is not covered by previous studies.

The following of this paper is organized as follows: In section 2, we give a brief introduction of current QA research lines and research on live text processing. In section 3, we describe how we constructed the dataset. In section 4, we give statistics of the dataset and analyse the timelineness and mathematical inference in the data. In section 5, we give evaluation results of baseline models and error analysis.

\section{Related Works}
In this section, we mainly introduce the various QA datasets which can be categorized as datasets with extractive answers, datasets with descriptive answers and datasets with multiple-choice questions.

\subsection{Datasets with Extractive Answers}
A number of QA datasets consist of numerous documents or passages which have considerable length. Each passage is equipped with several questions, answers of which are segments of the passage. The goal of a reading comprehension model is to find the correct text span. In other words, it may offer a begin position and an end position in the passage instead of generating the words itself. Such corpora are regarded as datasets with extractive answers.

The most famous dataset of this kind is Stanford Qustion Answering Dataset (SQuAD) ~\cite{rajpurkar2016squad}. SQuAD v1.0 consists of 107,785 question-answer pairs compiled by crowdworkers from 536 Wikipedia articles, and is much larger than previous manually labeled datasets. Over 50,000 unanswerable questions are added in SQuAD v2.0 ~\cite{rajpurkar2018know}. It is more challenging for existing models because they have to make more unreliable guesses. As performances on SQuAD have become a common way to evaluate models, some experts regard SQuAD as the ImageNet~\cite{Deng2009ImageNet} dataset in the NLP field. 

\begin{figure}[!t]                                                     %开始插图
\centering                                                                %图片居中
\includegraphics[scale=0.18]{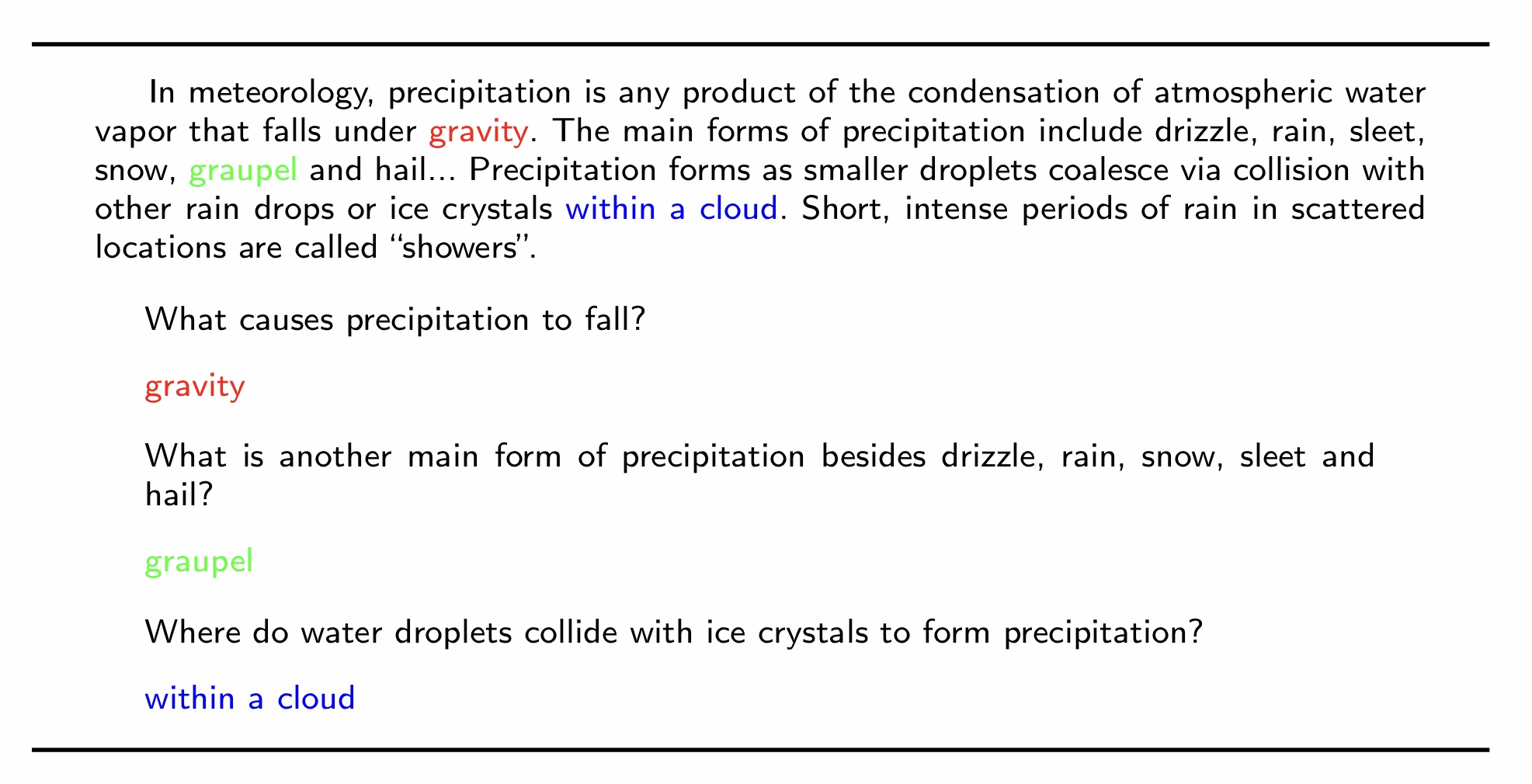}                                    %确定图片大小/位置
\caption{Examples of question-answer pairs in SQuAD}                                    %加名称和标签
\end{figure}

\begin{comment}
\begin{figure}[!t]                                                     %开始插图
\centering                                                                %图片居中
\includegraphics[scale=0.38]{squad1.jpg}                                    %确定图片大小/位置
\caption{SQuAD compared to other datasets}                                    %加名称和标签
\end{figure}
\end{comment}

Another frequently used dataset with extractive answers is CNN/Daily Mail dataset~\cite{Hermann2015Teaching}, which was released by Google DeepMind and University of Oxford in 2015. One shining point of it is that each entity is anonymised by using an abstract entity marker to prevent models from using word-level information or n-gram models to find the answer rather than comprehending the passage.

\begin{figure}[!t]                                                     %开始插图
\centering                                                                %图片居中
\includegraphics[scale=0.45]{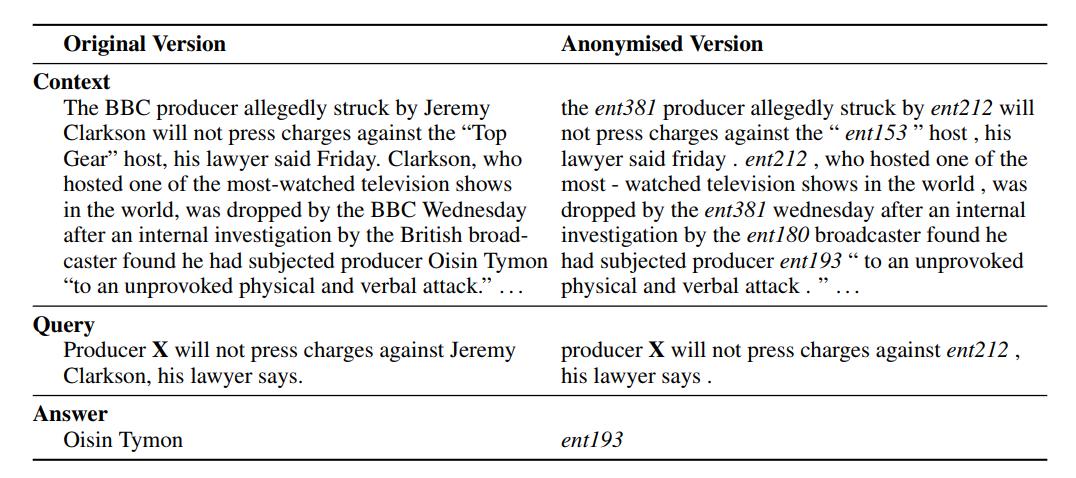}                                    %确定图片大小/位置
\caption{An example of anonymised entity in CNN/Daily Mail}                                    %加名称和标签
\end{figure}

\begin{CJK}{UTF8}{gbsn}
\begin{figure}
        \centering
       \begin{tabular}{p{120pt}p{120pt}p{50pt}p{50pt}}
       \hline
       \textbf{Text} & \textbf{Question} & \textbf{Choices}& \textbf{Answer}\\
         \hline
         
          ……\\
          \small{哈登弧顶控球！！面对克莱-汤普森紧逼！}\\
          \small{左侧横移！！！！}\\
          \small{哨响！克莱-汤普森逼得太紧了！吃到一次犯规！}\\
         \small{勇士要个长暂停！！！！！}\\
         &\small{停回来，勇士第一轮进攻能否得分？（罚球也算，直到球权转换）}&\small{能/不能} & \small{不能}\\
         \small{稍等！！！}\\
         \small{第一节还有7分29秒！}\\
         &\small{本节勇士队最后一分是否由伊戈达拉获得？}&\small{是/不是} & \small{不是}\\
         \small{好的！！！比赛继续！！！}\\
         \small{哈登走上罚球线！！！}\\
         \small{两罚都有！！14-9}\\
         \small{利文斯顿弧顶控球！！！}\\
         ……\\

        \hline
        \hline
       \end{tabular}
       \caption{A partial example of LiveQA timeline.}
       \label{liveex}
\end{figure}
\end{CJK}

CBT~\cite{Hill2015The}, NewsQA~\cite{Trischler2016NewsQA}, TriviaQA~\cite{Joshi2017TriviaQA} and many other datasets can also be categorized into this class. They  constitute a high proportion of MRC datasets, and can test the abilities of extractive models in various ways. The most closest work to ours is DROP~\cite{dua2019drop}, which is a QA dataset that requires discrete reasoning over the content of paragraphs. It requires the system to extract various pieces of numerical evidence and perform calculation on top of the numbers. Thus, we aim to construct a novel dataset, on which extractive models are likely to make mistakes in looking for the location of an answer, that the dataset can open a new research line for question answering by testifying the ability of models to understand timelineness.

% It will help to their strengthening, 
%%% 这句话什么意思？
% and it is the reason why we seek for a corpus with timelines.
%%%
\subsection{Datasets with Descriptive Answers}
Instead of selecting a span from the passage, datasets with descriptive answers require a reading comprehension model to generate whole and stand-alone sentences. These corpora are more closer to reality, because most questions in the real world cannot be solved simply by presenting a span or an entity. This kind of dataset is getting popular nowadays, and  may be the trend of the development of MRC datasets.\\ 
MS MARCO (Microsoft MAchine Reading Comprehension) ~\cite{Nguyen2016MS} is a dataset released by Microsoft in 2016. This dataset aims to address the questions and documents in the real world, as its questions are sampled from Bing's search query logs and its passages are extracted from web documents retrieved
by Bing. The questions in MS MARCO are about ten times as many as SQuAD, and each question is equipped with a human generated answer. The dataset also includes unanswaerable questions. All of the above characteristics make MS MARCO worthy of trying.

NarrariveQA~\cite{Ko2017The} is another dataset with descriptive answers released by DeepMind and University of Oxford in 2017. The dataset consists of stories, %which are 
books and movie scripts, with human written questions and answers based solely on human-generated abstractive summaries. Answering such questions requires readers to integrate information which may distribute across several statements throughout the document, and generate a cogent answer on the basis of this integrated information. In other words, they test that the reader comprehends language, not just that it can pattern match. We judge it a referential advantage of a dataset, so LiveQA requires the ability of tracking events as well as we show in Figure \ref{liveex}, which will be detailedly introduced in following sections.

\subsection{Datasets with Multiple-choice Questions}
Datasets with descriptive answers have various advantages, but they are relatively diﬃcult to evaluate the system performance precisely and objectively. Thus, corpuses with more gradable QA-pairs are also needed, which leads to the development of datasets with multiple-choice questions. Through diversified types of questions, these datasets can examine almost every ability of a reading comprehension model mentioned above and are easier to get a conclusive score.
Many datasets of this kind have been released in recent years, and they have covered multiple domains. For example, RACE~\cite{lai2017race} and CLOTH~\cite{xie2017large} are collected from English exams, MCTest~\cite{richardson2013mctest} is sampled from friction stories, and ARC~\cite{clark2018think} is extracted from science-related sentences. However, there is still not a reliable dataset which is built on sports events for MRC. Thus, our LiveQA dataset has the potential for filling several gaps in the field of MRC.

%%% 关于SQUAD和CNN的数据集的图或者去了，或者把它们和liveQA作比较

 \subsection{Live Text Processing}
 
 Previously, various studies have been conducted on automatically generate sports news from live text commentary scripts, which has been seen as a summarization task. Zhang et al.~\shortcite{zhang2016towards} proposed an investigation on summarization of sports news from live text commentary scripts, where they treat this task as a special kind of document summarization based on sentence extraction in a supervised learning to rank framework. Yao et al.~\shortcite{yao2017content} further verify the feasibility of a more challenging setting to generate news report on the fly by treating live text input as a stream for sentence selection. Wan et el.~\shortcite{10.1007/978-3-319-50496-4_80} studied dealing with the summarization task in Chinese. All these studies focuses on using the live text commentary scripts as the input of summarization and selecting sentences to form the summary. So far, we are the first to point out the importance of timelineness and mathematical reasoning in understanding live text commentary scripts.

\section{LiveQA: Dataset Construction}
In this section we introduce how to construct LiveQA from the raw Hupu text and present the corpus statistics.
The whole process of building LiveQA mainly includes  crawling the raw data and acquiring the game texts with corresponding quizzes.
%%% 这儿是否可以补一段数据如何存储的？
\subsection{Data Crawling}
In Hupu, each game has a unique ID which is connected with its url. We collected the IDs from the Hupu's live schedule pages. Their formats are {\em https://nba.hupu.com/games/year-month-date}. There are links to all the NBA games so that their IDs can be saved. After we saved the IDs into a file, we used the web debugging tool Fiddler to get a sample of the url of a game, and then changed the IDs in the url to make access to all the games. We are authorized by the legal department of hupu website to construct the dataset for only academical purpose.
%%% 关于爬取hupu数据的版权说明，这儿最好有个解释。

\subsection{Data Processing}
%The passages and their questions  are usually sepad in most previous datasets.
 Most previous datasets usually do not care for the storing positions of the passages and their questions.
 But in our dataset, the quizzes and the contexts shouldn't be separated because the time (the position) one quiz occurs is quite important for the final answer. If we separate the quizzes and their contexts, 
 most quizzes may have different answers and cannot be answered even by human.
% take them out into a separate document, most of them will be impossible to be answered even by human.  
 Here we use some rules to clean the dataset. 
 The lines starting with '@' are always interactions between the host and some active readers, which are irrelevant to the game. During the half-time break, the host will give out some "gift" questions to please the readers waiting for the second-half. 
 Some of the questions appear like normal quizzes, but they need information outside the game to answer them, thus we exclude them from the data (i.e.. which team won more matches in the history?). Usually they have a prefix -- \begin{CJK}{UTF8}{gbsn}"中场福利”\end{CJK}  in common. Besides, we exclude the descriptions of pictures from our data. 
 %%% 这儿最好给出一个实际的文本例子和相关问题。由于gift questions本身有一些猜的因素，这儿能否展开说一下抽取过程中的困难。这儿最好能多展开说说。
 
 \subsection{Data Structure}
 
 Here we give an explanation for the structure for each independent data sample.
 
 For each live-stream of one match, the timeline data is sorted in time order, where the questions are inserted into the corresponding timeline position so that the timeline features of the questions could be inferred. As we show in Figure \ref{liveex}, the plain content text and the question text share the same timeline, but question records have choices and answers along with the text. For each record in the timeline, it either contains a piece of live-stream text or a question bonded with the corresponding choices and the correct answer.
 Each question has two answer choices.

\section{Dataset Statistics}

\begin{table}[h]
\begin{center}
\begin{tabular}{ll}
\hline
\bf Element                              & \bf Count        \\
\hline
Document                          & 1,670 \\
Sentences in Total & 1786616\\
Sentences in Average & 1069.83\\
Quizzes in Total& 117050\\
Quizzes in Average                         & 70.09 \\

\hline
\end{tabular}
\end{center}
\caption{\label{stat} The details of statistics of the dataset.}
\end{table}

We show the statistics of the dataset in Table \ref{stat}. The LiveQA dataset contains 1,670 documents, each of which has 70.09 quizzes and 1069.83 sentences on average. 
Next we analyze the questions from two different views.
First, we simply classify the questions according to the positions of their answers. 
%The first criterion is the location we can find one's answer. 
In general, some of the questions can be solved by extracting information from neighboring sentences, which involves a time period of the origin game. Such questions occupy
68.6\% of all the questions. Some questions can be replied only by summarizing all the information after the game ends and occupy about 30.6\%. Still, there exists a small percentage (0.8\%) of questions which are impossible to be answered from the passage.
Table \ref{loc} lists some examples for each type of questions.

% \begin{table}[h]
% \begin{center}
% \begin{tabular}{ll}
% \hline
% \bf Element                              & \bf Count        \\
% \hline
% Time-period                          & 68.6\% \\
% Full game & 30.6\%\\
% No answer & 0.8\%\\

% \hline
% \end{tabular}
% \end{center}
% \caption{\label{stat} The details of question types.}
% \end{table}

%We classify and describe the questions in two criteria. The first criterion is the location we can find one's answer. In general, Some of the questions can be solved by extracting information from a certain sentence near it. Some others can be replied by summarizing information from events happening right after the quiz (but not the whole passage). While others can only be answered after the game ends. Few of them are impossible to be answered.  
\begin{CJK}{UTF8}{gbsn}
\begin{table*}[t!]
\centering
\begin{tabular}{|p{2cm}|p{1.5cm}|p{4cm}|p{4.5cm}|}
%{|c|c|c|c|}
\hline
Question type&Proportion&Example&Translation\\
\hline
\centering
\multirow{2}[6]{2cm}{Answered after the game ends}&
\centering
\multirow{2}[6]{1.5cm}{30.6\%}&
本场森林狼能否赢快船4分或更多？&Will the Timberwolves beat the Clippers by more than 4 points?\\
\cline{3-4}
& &本场比赛谁会赢？&Which team will win?\\
\cline{3-4}
\hline
\multirow{2}[8]{2cm}{Answered through the context}&
\centering
\multirow{2}[8]{1.5cm}{68.6\%}&
第二节谁先命中三分球？&Which team will make a three-pointer first in the second quarter?\\
\cline{3-4}
& &首节最后一分会不会由罚球获得？&Will the last point in the first quarter scored through a free-throw?\\
\cline{3-4}
\hline
\multirow{2}[9]{2cm}{Impossible to answer}&
\centering
\multirow{2}[9]{1.5cm}{0.8\%}&
第二节比赛开始1分30秒时间内会不会有三分球命中？&Will a three-pointer be made in the first 90s of the second quarter?\\
\cline{3-4}
& &本场比赛会不会在北京时间10时58分之前结束？&Will the game end before 10:58 a.m.?\\
\cline{3-4}
\hline
\end{tabular}
\caption{Questions statistics and examples sorted by the location of their corresponding evidence.}
\label{loc}
\end{table*}
\end{CJK}

\begin{CJK}{UTF8}{gbsn}
\begin{table*}[t!]
\centering
\begin{tabular}{|p{2cm}|p{1.5cm}|p{4cm}|p{4.5cm}|}
%{|c|c|c|c|}
\hline
Question type&Proportion&Example&Translation\\
\hline
\centering
\multirow{2}[6]{2cm}{Comparison}&
\centering
\multirow{2}[6]{1.5cm}{16.6\%}&
勒布朗-詹姆斯本场能否得到26分或更多？&Will Lebron James get 26 points or more in this game?\\
\cline{3-4}
& &本场谁的得分会更高？&Who will get higher score in this game?\\
\cline{3-4}
\hline
\multirow{2}[9]{2cm}{Calculation}&
\centering
\multirow{2}[9]{1.5cm}{25.4\%}&
本场凯尔特人能否赢猛龙3分或更多？&Will the Celtics beat the Raptors by more than 3 points?\\
\cline{3-4}
& &本场两队总得分能否达到207分或更多？&Will the total score of the two teams reach 207 points or more?\\
\cline{3-4}
\hline
\multirow{2}[10]{2cm}{Inference}&
\centering
\multirow{2}[10]{1.5cm}{28.5\%}&
暂停回来，雷霆队首次进攻能否得分？&After the timeout, will the Thunder score in their first round of attack?\\
\cline{3-4}
& &第二节比赛雷霆队最后一分会不会由威斯布鲁克得到？&Will the last point of the Thunder in the second quarter be got by Westbrook?\\
\cline{3-4}
\hline
\multirow{2}[8]{2cm}{Tracking}&
\centering
\multirow{2}[8]{1.5cm}{29.5\%}&
太阳队能否在本场命中8个或更多三分球？&Will the Suns make 8 three-pointers or more in this game?\\
\cline{3-4}
& &凯文-乐福首节犯规数会不会达到2次？&Will Kevin Love commit 2 fouls or more in the first quarter?\\
\cline{3-4}
\hline
\end{tabular}
\caption{Questions statistics and examples sorted by how the inference process is done.}
\label{inf}
\end{table*}
\end{CJK}

Because most of the questions are associated with some numerical data in the game, we also classify the questions according to how the numerical data is performed. Four types of operations are commonly used including: \textit{Comparison}, \textit{Calculation}, \textit{Inference} and \textit{Tracking}. Then the questions are correspondingly classified are introduced in the following subsections. We also give some examples in Table \ref{inf}.

%we propose the second classifying criteria and judge a question depending on how the numerical data will be processed to get the answer. In the following subsections, we will introduce four types of questions in our dataset under this criteria. 
%The result of classification is shown in Table \ref{inf}. 

\subsection{Comparison}
To answer the comparison questions, we usually need to find the comparative figures for the corresponding objects.
For example, the commentator asks which of the two players will score more or which team will win.
The second row in Table \ref{inf} belong to the \textit{Comparison} questions. The easiest way to solve this kind of questions is to find the two figures appearing in the text and comparing them. 
It is likely to acquire such figures after the game ends, and the specific figures usually appear together in a summary of the game in the end. Thus, matching techniques are still necessary to the final answer.

%The first two questions in Table \ref{inf} belong to the \textit{Comparison} questions. The easiest way to solve this kind of questions is simply comparing two figures appearing in the text. They are more likely to require information given after the game ends. Most of them compare very specific figures and the figures are usually mentioned together and clearly. For example, the commentator asks which of the two players will score more or which team will win, as both will appear in a summary of the game appearing in the end. So matching may make the answers easy to access.

\subsection{Calculation}
The \textit{Calculation} questions require extracting two or three figures and calculating their sum or difference. They differ from the Comparison questions in two ways -- the figures are more scattered and a calculation step is needed. This means that a respondent has to look for more information efficiently. After the figures are obtained, if a respondent misjudges the type or the direction of the calculation, he will still probably get a wrong answer. 
Similar to the \textit{Comparison} questions, the \textit{Calculation} questions are mainly dependent on the correct sentences where the figures are located. These two kinds of questions are relatively easy compared to those ones which are not based on certain sentences. The second row in Figure \ref{inf} give two example questions.
%However, it remains in the group of less challenging questions because if a respondent finds the correct sentences where the information is located, he will be likely to solve it through some simple methods, while there are some others which are hard to get a certain sentence that a question bases on. 
\subsection{Inference}
The third and fourth type of questions require the ability of summarizing and tracking information. A question of the third type needs a respondent to infer some figures through the text. For example, a question may be "After this timeout, will the Cavaliers score in the first round of attack?". The commentator  obviously will not say that "The Cavaliers scored 2 points." or "The Cavaliers didn't score." A respondent may get the answer as "JR Smith makes a 2-point shot." Another example is "Will the last point of this quarter be scored through a free throw?" The information comes from the text of "Anthony Davis makes his second free throw ... The match ends!". It is impossible to get a reasonable answer by matching. 

\subsection{Tracking}
The Tracking questions require more scattered information. A respondent should collect and accumulate specific information from a part of the passage, as the question is based on events happening repeatedly in a quarter or half of the game. For example, some questions ask about how many free-throws a player \textit{A} will make in a quarter. As this figure does not appear in the passage, a respondent needs to count how many times the event '\textit{A} makes a free-throw' occurs. In other words, it is necessary to track events relevant to the player '\textit{A}' and 'free-throw'. When the player(\textit{A}) is replaced with one team name, the new question is even 
more difficult because the information about each player belonging to the team should be tracked. Therefore, information tracking leads this kind of questions to be the most challenging ones in the dataset.

\section{Baseline Models and Results}
\subsection{Models}
To evaluate the QA performance on the LiveQA dataset, we implement 3 baseline models. The first is based on  random selection, where the system randomly chooses a choice as the answer. The second is to choose the dominant option of each question. More concretely, 80.0\% of questions are in format of 'yes' and 'no', where 57.8\% has the answer 'no'. For the other multiple choice questions, 50.6\% of them take the second option as the right answer.
%the answer to 57.8\% of which is 'no', while 50.6\% of the other questions take the second option as their right answer.
%Thus, if the two options are in format of 'yes' and 'no', we choose 'no', otherwise we choose the second option.
Thus, for 'yes/no' questions, we choose 'no', otherwise we choose the second option.

We also build a neural-network style baseline for our dataset to evaluate how state-of-the-art QA systems perform on the LiveQA dataset. Due to the uniqueness of our dataset, most of existing machine comprehension models are not suitable to it. For example, the QANet \cite{yu2018qanet} model, which used to be a state-of-art model of SQuAD \cite{rajpurkar2016squad}, is unavailable because it predicts the probability distribution of an answer's starting position and ending position in the context. But in LiveQA, a number of right answers do not directly appear in the context (e.g. an answer in format of 'can' or 'cannot'). 
Up to now, none of machine reading comprehension models has been designed for a dataset with consideration of timeline and mathematical computations. 
That means that the existing ones will not be likely to perform well on our dataset. The closest work to ours is multi-hop question answering, and thus we use a novel model Gated-Attention Reader \cite{dhingra2016gated} to experiment on LiveQA. 

Gated-Attention Reader (GA) is an attention mechanism which uses multiplicative interactions between the query embedding and intermediate states of a recurrent neural network reader. GA enables a model to scan one document and the questions iteratively for multiple passes, and thus the multi-hop structure can target on most relevant parts of the document. It used to be the state-of-art model of several datasets, such as CNN/Daily Mail dataset \cite{Hermann2015Teaching} and CBT dataset \cite{hill2015goldilocks}.

The full context, which is usually composed of more than 1,000 sentences on average, is too heavy for GA as input. To apply GA to our dataset, we propose a pipeline method to first extract %the answer from
a set of candidate evidence sentences from the full content, and then apply the GA model on this set of sentences to predict the final answer. We employ TF-IDF style matching score to extract 50 most relevant sentences as the supporting evidence. %a question's document.
To improve the accuracy of selecting the evidence candidates, if the question clearly requires some information after the game ends, we use the ending part of the content as the input.
%%% 这儿不知道是否可以具体一些？

Specifically, taken the embedding representation of a token, the Bi-directional Gated Recurrent Units (BiGRU) process the sequence in both forward and backward directions to produce two sequences of token-level representations, which are concatenated at the output as the final representation of the token. 
To perform multi-hop inference, the GA model reads the document and the query over $k$ horizontal layers, where layer $k$ receives the contextual embeddings $X_{(k-1)}$ of the document from the previous layer. 
At each layer, the document representation $D^{(k)}$ is computed by taking the full output of a document BiGRU where the previous layer embedding $X_{(k-1)}$ is the input. At the same time, a layer-specific query representation $Q^{(k)}$ is computed as the full output of a separate query BiGRU taking the query embedding $Y$ as the input. The Gated-Attention is applied to $D^{(k)}$ and $Q^{(k)}$ to compute the contextual embedding $X^{(k)}$.

\begin{equation}
    X^{(k)} = GAttn(BiGRU(X^{(k-1)}), BiGRU(Y))
\end{equation}

After obtaining the query-awared document representation, we perform answer prediction by matching the similarity of answer and content. We use bidirectional Gated Recurrent Units to encode the candidate answers into vectors $A^(i)$, and then we compute matching score between summarized document and candidates using a bilinear attention.
Finally we calculate the probability distribution of the options with softmax. The operations are similar to those in RACE \cite{lai2017race}.

\begin{equation}
    s = softmax([Blin(A^i,D^{(k)});]^{i=1}_{n})
\end{equation}

\subsection{Model Evaluation}

\begin{table}[h]
\begin{center}
\begin{tabular}{ll}
\hline
\bf Model                              & \bf Acc        \\
\hline
Random                          & 50.0\% \\
Dominant & \textbf{56.4\%}\\
GA & 53.1\%\\

\hline
\end{tabular}
\end{center}
\caption{\label{rel} The results of different baseline models on the test set. Random denotes randomly selecting an answer. Dominate denotes selecting the dominate option. GA denotes the gated-attention reader.}
\end{table}

For the three baseline models, performance is reported with the accuracy on the test set in Table \ref{rel}. The random selection method (Random) scores 50.0\%, while the dominant option method (Dominate) reaches a score of 56.4\%, which shows that our dataset does not have a certain pattern for the answers. Meanwhile, GA, which is a strong baseline for previous question answering problems, failed to perform better than the dominant option method and only achieves a score of 53.1\%. Such results show that our dataset is challenging and needs further investigation for model design. In future work, how to incorporate temporal information and mathematical calculation into a QA model is the focus.
%a model should consider the timeliness and the mathematical calculation to achieve better performance.

\subsection{Case Study}

\begin{CJK}{UTF8}{gbsn}
\begin{table*}[ht]
\centering
\begin{tabular}{|p{3.5cm}|p{3.5cm}|p{2.5cm}|p{2.5cm}|}
%{|c|c|c|c|}
\hline
Question&Translation&Correct answer&Answer given by the model\\
\hline
跳球之争！本场比赛哪支球队获得第一轮进攻球权？&Jump ball fight! Which team will win the chance of the first round of offence?&勇士(The Warriors)&勇士(The Warriors)\\
\hline
湖人全场总得分是奇数还是偶数？&Will the total score of the Lakers at the end of the game be odd or even?&奇数(odd)&奇数(odd)\\
\hline
尼克杨第二节能否命中3分球？&Can Nick Young make a three pointer in the second quarter?&能(Yes)&不能(No)\\
\hline
第三节结束，76人能否领先湖人4分或更多？&At the end of the third quarter, Will the 76ers lead the Lakers by 4 points of more?&不能(No)&能(Yes)\\
\hline
谁先获得30分？&Who will score his 30th point earlier?&24分的哈登(James Harden who has got 24 points)&25分的托马斯(Isaiah Thomas who has got 25 points)\\
\hline
\end{tabular}
\caption{Cases in the experimental results}
\label{case}
\end{table*}
\end{CJK}

In this subsection, we further analyze the prediction ability of the GA model.
Table \ref{case} shows some prediction cases in experimental results. 
From the first two questions, we can see that the model gives the correct answers when judging the result of a specific event. But for the other three questions which involve multiple events, the model fails to answer them correctly. A possible explanation is that, although GA is designed for multi-hop inference, it lacks ability in both  information tracking and math calculation, which makes it difficult for the model to track down some complicated events. 

We can see, for reading comprehension models that extract answers based on the similarity between the answer and the content, they would fail on LiveQA due to the fact that they cannot track down temporal information nor perform mathematical calculation.
To outperform existing models on LiveQA, the system should consider focusing on tracking information of a certain event through the timeline. It should also have the ability to perform mathematical inference between different contents. 
%And we guess same phenomenon may appears in other comprehension models, especially the ones producing extractive answers, for similarity between candidates and their absence in document. To outperform on LiveQA, an architecture should be able to focus on figures and cast attention to decentralized parts of document. We expect new models to overcome the challenging problems in the future. 

\section{Conclusion}
In this paper, we present LiveQA, a question answering dataset constructed from play-by-play live broadcast. LiveQA can evaluate a machine reading comprehension model in its ability to understand the timeline, track events and do mathematical calculation. It  consists of 117k questions, which are time-dependent and need math inference. Due to the novel characteristics, it is hard for existing QA models to perform well on LiveQA. We expect our dataset will stimulate the development of more advanced machine comprehension models.

\section*{Acknowledgement}
We thank the anonymous reviewers for their helpful comments on this paper. This work was partially supported by  National Natural Science Foundation Project of China (61876009), National Key Research and Development Project
(2019YFB1704002), and National Social Science Foundation Project of China (18ZDA295). The corresponding author of this paper is Sujian Li.

\bibliographystyle{ccl}
\bibliography{ccl2020-en}

\end{document}